# A Dissipative Particle Swarm Optimization

Xiao-Feng Xie, Wen-Jun Zhang, Zhi-Lian Yang

Institute of Microelectronics, Tsinghua University, Beijing 100084, P. R. China

xiexiaofeng@tsinghua.org.cn, zwj@mail.tsinghua.edu.cn, yangzl@tsinghua.edu.cn

**Abstract** - A dissipative particle swarm optimization is developed according to the self-organization of dissipative structure. The negative entropy is introduced to construct an opening dissipative system that is far-from-equilibrium so as to driving the irreversible evolution process with better fitness. The testing of two multimodal functions indicates it improves the performance effectively.

## 1. Introduction

Particle swarm optimization (PSO) is an evolutionary computation technique developed by Kennedy and Eberhart in 1995 [1, 2]. The underlying motivation for the development of PSO algorithm was social behavior of animals such as bird flocking, fish schooling, and swarm theory [3]. One of reasons that PSO is attractive is that there are very few parameters to adjust. Work presented in [4, 5] describes the complex task of parameter selection in the PSO model. Several researchers have analyzed the performance of the PSO with different settings, e.g., neighborhood settings [6], cluster analysis [7], etc. It has been used for approaches that can be used across a wide range of applications [8].

However, studies by Angeline [9] showed that although PSO discovered reasonable quality solutions much faster than other evolutionary algorithms, it did not possess the ability to improve upon the quality of the solutions as the number of generations was increased. This indicates although the current simple social model in PSO suggests the irreversible process toward higher fitness by the weak selection that recording best historical experience, it is lacking of enough capability to "sustainable development" (i.e. get better fitness as evolution process goes on).

Understanding the emergence and evolution of biological and social order has been a fundamental goal of evolutionary theory. Current rapid development of methods of complex systems dynamics [10-13] argue that order can only be maintained by *self-organization*. Structures of increasing complexity in open systems based on energy exchanges with the environment were developed into a general thermodynamic concept of *dissipative structures* by Prigogine [10]. With the selection that providing the direction of evolution, the self-organization of dissipative systems interprets the general phenomenon of a nonequilibrium system evolving to an order state as a result of fluctuations.

Self-organizing dissipative systems allow adaptation to the prevailing environment, i.e. they react to changes in the environment with a thermodynamic response which makes the systems extraordinarily flexible and robust against perturbation of the outer conditions. An entirely new technology will have to be developed to tap the high guidance and regulation potential of self-organizing systems for technical processes. The superiority of self-organizing systems is illustrated by biological and social systems where complex products can be formed with unsurpassed accuracy, efficiency and speed.

This paper describes a variant of particle swarm, termed *dissipative* PSO, which introduces negative entropy to stimulate the model in PSO operating as a dissipative structure. Both standard and dissipative versions are compared on two multimodal optimization problems typically used in evolutionary optimization research. The results show that the additional fluctuations supply some advantage to particle swarm on " sustainable development" .

## 2. Standard particle swarm optimization (SPSO)

The fundament to the development of PSO is a hypothesis [14] that social sharing of information among conspeciates offers an evolutionary advantage. PSO is similar to the other evolutionary algorithms in that the system is initialized with a population of random solutions. However, each potential solution is also assigned a randomized velocity, and the potential solutions, call *particles*, corresponding to individuals. Each particle in PSO flies in the D-dimensional problem space with a velocity which is dynamically adjusted according to the flying experiences of its own and its colleagues. The location of the *i*th particle is represented as $X_i = (x_{i1}, \ldots, x_{id}, \ldots, x_{iD})$, where $x_{id} \in [l_d, u_d]$, $d \in [1, D]$, $l_d$, $u_d$ are the lower and upper bounds for the *d*th dimension, respectively. The best previous position (which giving the best fitness value) of the *i*th particle is recorded and represented as $P_i = (p_{i1}, \ldots, p_{id}, \ldots, p_{iD})$, which is also called *pbest*. The index of the best particle among all the particles in the population is represented by the symbol *g*. The location $P_g$ is also called *gbest*. The velocity for the *i*th particle is represented as $V_i = (v_{i1}, \ldots, v_{id}, \ldots, v_{iD})$, is clamped to a maximum velocity $V_{max} = (v_{max,1}, \ldots, v_{max,d}, \ldots, v_{max,D})$, which is specified by the user.

The particle swarm optimization concept consists of, at each time step, changing the velocity and location of each particle toward its *pbest* and *gbest* locations according to the equations (1a) and (1b), respectively:

$$v_{id} = w * v_{id} + c_1 * rand() * (p_{id} - x_{id})$$
$$+ c_2 * rand() * (p_{gd} - x_{id}) \quad (1a)$$
$$x_{id} = x_{id} + v_{id} \quad (1b)$$

Where $w$ is inertia weight [15], $c_1$ and $c_2$ are acceleration constants [8], and *rand*() is a random function in the range [0, 1]. For equation (1a), the first part represents the inertia of pervious velocity; the second part is the " cognition" part, which represents the private thinking by itself; the third part is the "social" part, which represents the cooperation among the particles [16]. If the sum of accelerations would cause the velocity $v_{id}$ on that dimension to exceed $v_{max,d}$, then $v_{id}$ is limited to $v_{max,d}$. $V_{max}$ determines the resolution with which regions between the present position and the target position are searched [4, 8].

The process for implementing PSO is as follows:

a). Initialize a population (array) which including *m* particles, For the *i*th particle, it has random location $X_i$ in the problem space and for the *d*th dimension of velocity $V_i$, $v_{id} = Rand_2() * v_{max,d}$, where $Rand_2()$ is in the range [-1, 1];

b). Evaluate the desired optimization fitness function for each particle;

c). Compare the evaluated fitness value of each particle with its *pbest*. If current value is better than *pbest*, then set the current location as the *pbest* location. Furthermore, if current value is better than *gbest*, then reset *gbest* to the current index in particle array;

d). Change the velocity and location of the particle according to the equations (1a) and (1b), respectively;

e). Loop to step b) until a stop criterion is met, usually a sufficiently good fitness value or a predefined maximum number of generations $G_{max}$.

The parameters of standard PSO includes: number of particles *m*, inertia weight *w*, acceleration constants $c_1$ and $c_2$, maximum velocity $V_{max}$.

**3. Dissipative particle swarm optimization (DPSO)**

3.1 Self-organization of dissipative structures

Three realms of thermodynamics are differentiated by Prigogine [10, 12]. In equilibrium realm, it has maximal entropy. Close to equilibrium realm, where the rates of processes are linear functions of the underlying forces, systems evolve toward a stationary equilibrium state characterized by the minimum of entropy production compatible with the boundary conditions. In the third, far-from-equilibrium realm of thermodynamics, with the nonlinearity of flows and forces, system leaves the unstable state and evolves to one of the many possible new states. These new states can be highly organized states. The features of far-from-equilibrium systems imply that initial conditions and random fluctuations may have a permanent effect on the system's development. Since the creation of organized nonequilibrium states are due to dissipative processes, they are called dissipative structures.

The self-organization of dissipative structure is frequently used as a generic dynamic concept to describe the evolution of nonlinear systems [11, 12, 17]. Often, such applications do not even refer to the thermodynamic foundations, but far-from-equilibrium conditions are taken for granted as a prerequisite for developing increasingly complex structures in evolutionary processes.

Moreover, self-organization requires a system consisting of multiple elements in which nonlinear relations of feedback between system elements are present [13, 17]. Positive feedback is necessary for the amplifying of random fluctuations so as to drive the dissipative system into an order state distinguishable from the random configuration of thermodynamic equilibrium. In order to maintain the order state, some negative feedback is also present that dampen the effects of further fluctuations. Self-organization thus results from interplay of positive and negative feedback, which of these is actually realized depends on random fluctuations. Therefore, system development is permanently affected by random fluctuations. Accordingly, the Prigogine school refers to the self-organization of dissipative systems as "order through fluctuations" [11, 12].

With the spatiotemporal symmetry broking by selection for higher fitness, the self-organization of dissipative structure provides the inevitability of the general phenomenon of increasingly complex structures in a nonequilibrium system out of chanciness.

### 3.2 Social model in standard PSO

The simple social model in standard PSO has some characteristics for self-organization of dissipative structure. The selection of keeping best historical experience constructs the irreversible process toward higher fitness. Then the process starts as the initialization of particles with random locations and velocities bring the system far-from-equilibrium. The randomicity of fluctuations is provided by the *rand*() function for the acceleration constants. The "social" part of equation (1a) ensures non-linear relations of positive and negative feedback between particles according to the cooperation and challenge with the particle with best experience so far.

However, these characteristics that similar to dissipative structure may be faded as evolution process goes on.

Firstly, as the function of successive negative feedback by imitating the best particle among all the particles (*gbest*), the best historical experience of every particle (*pbest*) are apt to be similarly, which means the "social" part is tend to be ineffective, and the swarm is inclined to be decomposed as independent particles which lost nonlinear relations of feedback between particles.

Secondly, the swarm may be damped to equilibrium state. In order to solve different problems, the concept of inertia weight $w$ was introduced by Shi [15] to satisfy the requirements for different balances between the local search ability and global search ability, i.e. to be in equilibrium or in chaos. Since the chaotic state should be avoided to accelerate the evolution process, the small or time decreasing $w$ is usually adopted [4], which will diminish the diversity of swarm and lead to equilibrium.

For an extreme case, if the particles have same locations, same *pbest*s, and all in zero velocities at certain evolution stage (for example, initialization stage), then the swarm is in stationary equilibrium with no possibility to evolution.

### 3.3 Dissipative particle swarm optimization

If the swarm is going to be in equilibrium, the evolution process will be stagnated as time goes on. To prevent the trend, an opening dissipative system DPSO is constructed by introduces negative entropy through additional chaos for particles, with the following equations (2a) and (2b) that is executed after equation (1) in the step d) of SPSO.

The chaos for velocity of particle is represented as:

**IF** ($rand() < c_v$) **THEN** $v_{id} = rand()*v_{max,d}$ (2a)

The chaos for location of particle is represented as:

**IF** ($rand() < c_l$) **THEN** $x_{id} = Random(l_d, u_d)$ (2b)

Where $c_v$ and $c_l$ are chaotic factors that in the range [0, 1], When $Random(l_d, u_d)$ is a random value between $l_d$ and $u_d$.

As in an opening system, the flying of a particle is not only referring to the historical experiences, but also effected by environment. The chaos introduces the negative entropy from outer environment, which will keep the system in far-from-equilibrium state. Then the self-organization of dissipative structure comes into being with the inherent nonlinear interactions in swarm and leads to "sustainable development" from the fluctuations.

This dissipative PSO model can be mapping into human social creative activity for exploring new knowledge. People accept new information from the environment frequently and get rid of general experiences consciously, found fresh knowledge space which is far from old and general one, carry through various nonlinear information sharing and competition among social members, so that new rudiment of thinking info will appear and be magnified, new knowledge then comes into being.

## 4. Experimental setting

In order to test the capability of the dissipative PSO to "sustainable development", two multimodal functions that are commonly used in the evolutionary computation literature [5, 9] are used. Both functions are designed to have minima at the origin.

The function $f_1$ is the generalized Rastrigrin function:

$$f_1 = \sum_{d=1}^{D} (x_d^2 - 10\cos(2\pi x_d) + 10) \quad (3a)$$

The function $f_2$ is the generalized Griewank function:

$$f_2 = \frac{1}{4000}\sum_{d=1}^{D} x_d^2 - \prod_{d=1}^{D}\cos\left(\frac{x_d}{\sqrt{d}}\right) + 1 \quad (3b)$$

For $d$th dimension, $x_{max,d}=10$ for $f_1$; $x_{max,d}=600$ for $f_2$. For both functions, the initialization range $x_d \in [-x_{max,d}, x_{max,d}]$ (for equation (2b), $x_{id} = Random(-x_{max,d}, x_{max,d})$), maximum velocity $v_{max,d}=x_{max,d}$. Acceleration constants $c_1=c_2=2$. The fitness value is set as function value. For all the figures that mentioned, the number of particles $m$ is fixed at 20. We had 500 trial runs for every instance.

## 5. Results and discussion

Figure 1, 2 and figure 3, 4 shows the mean fitness value of the best particle found with different $w$, $c_v$ and $c_l$ for the Rastrigrin and Griewank function, respectively. Where $c_v=c_l=0$, i.e. no additional negative entropy, means the standard PSO version, inertia weights $w$ are varied from 0 to 1, $c_v$ and $c_l$ are set as 0.001 and 0.002 for one parameter and as 0 for another parameter to test different status, $G_{max}$ is set as 1000 and 1500 generations corresponding to the dimensions 10 and 20, respectively.

By looking at the shape of curves in these figures, it is easy to see a "balance point" for SPSO, i.e. a value of $w$ with best mean fitness, which indicating a balance between the local and global search ability. When the $w$ is larger than the balance value, the SPSO is going to be in chaotic state which lacking of local search ability. There has almost no difference for the performance between the standard and the dissipative PSO version. When the $w$ is less than the balance value, the SPSO is apt to be in equilibrium state which lacking of global search ability. With the introduced negative entropy, both cases of the dissipative PSO version show excellent performance than the standard PSO version. Moreover, the chaos for location seems more effective than for velocity since it introduces large fluctuations, and is not affected by the value of inertia weight directly.

In addition, for standard PSO, it can be found an interesting phenomenon that is not according with the original anticipation [15] of decreasing performance with decreasing global search ability when $w$ is decreasing to zero, which indicates some unclear mechanisms may exists in the variation of $w$ that should be studied in the future.

In order to investigate whether the dissipative PSO scales well or not, different numbers of particles $m$ are used for each function which different dimensions. The numbers of particles $m$ are 20, 40, 80 and 160. $G_{max}$ is set as 1000, 1500 and 2000 generations corresponding to the dimensions 10, 20 and 30, respectively. Table 1 gives the additional test conditions, where $SF_0$ is the results by Shi and Eberhart [5] with an asymmetric initialization method and a linearly decreasing $w$ which from 0.9 to 0.4. $SF_1$ provides a transitional comparison to $SF_0$ as a symmetric initialization in this work. $DF_2$ and $DF_3$ are DPSO with $c_v=0$, $c_l=0.001$, and the $w$ of $DF_3$ is fixed at 0.4.

Table 2 and 3 lists the mean fitness value of the best particle found for the Rastrigrin and Griewank function, respectively.

The little difference of the results between $SF_0$ and $SF$ verifies that PSO were only slightly affected by the asymmetric initialization [9]. With same setting with linearly decreasing $w$, $DF_2$ is superior to $SF_0$ for Rastrigrin function, and is similar to $SF_0$ for Griewank function. However, for $DF_3$, which $w$ is fixed as 0.4, it shows overwhelming superiority to $SF_0$ for Rastrigrin function, and is also superior to $SF_0$ in most cases for Griewank function. The results suggest the performance can be improved by introduce negative entropy into the dissipative system as $w$ is small.

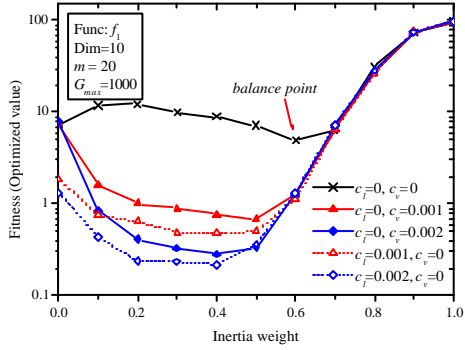

FIG. 1 10-D Rastrigrin function with different $c_v$ and $c_l$

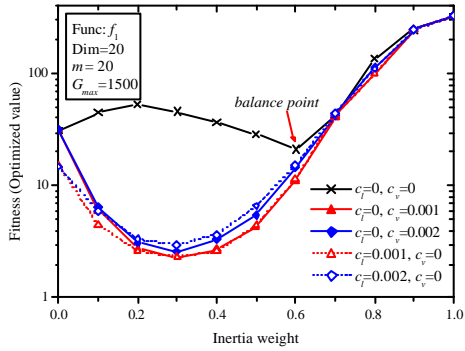

FIG. 2 20-D Rastrigrin function with different $c_v$ and $c_l$

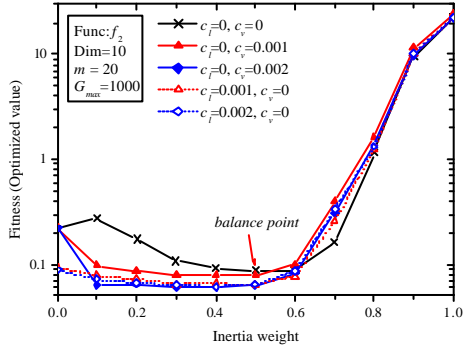

FIG. 3 10-D Griewank function with different $c_v$ and $c_l$

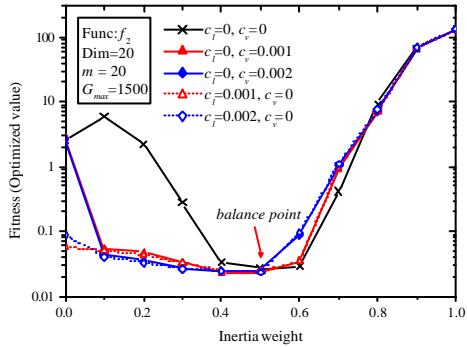

FIG. 4 20-D Griewank function with different $c_v$ and $c_l$

TABLE 1: Test conditions for standard and dissipative PSO

| Type | $SF_0$[5] | $SF_1$ | $DF_2$ | $DF_3$ |
|---|---|---|---|---|
| PSO version | SPSO | SPSO | DPSO | DPSO |
| Initialization | asymmetric | symmetric | symmetric | symmetric |
| Inertia weight | 0.9 to 0.4 | 0.9 to 0.4 | 0.9 to 0.4 | 0.4 |

TABLE 2: The mean fitness values for the Rastrigrin function

| $m$ | Dim. | $G_{max}$ | $SF_0$[5] | $SF_1$ | $DF_2$ | $DF_3$ |
|---|---|---|---|---|---|---|
| 20 | 10 | 1000 | 5.5572 | 5.20620 | 3.08128 | 0.47068 |
|  | 20 | 1500 | 22.8892 | 22.77236 | 13.85226 | 2.57289 |
|  | 30 | 2000 | 47.2941 | 49.29417 | 33.11479 | 7.32582 |
| 40 | 10 | 1000 | 3.5623 | 3.56974 | 1.62999 | 0.07619 |
|  | 20 | 1500 | 16.3504 | 17.29751 | 10.37524 | 1.30880 |
|  | 30 | 2000 | 38.5250 | 38.91423 | 24.8981 | 6.21067 |
| 80 | 10 | 1000 | 2.5379 | 2.38352 | 0.71879 | 0.00796 |
|  | 20 | 1500 | 13.4263 | 12.90195 | 7.25417 | 0.74955 |
|  | 30 | 2000 | 29.3063 | 30.03748 | 19.31247 | 4.22646 |
| 160 | 10 | 1000 | 1.4943 | 1.44181 | 0.22699 | 0.00199 |
|  | 20 | 1500 | 10.3696 | 10.04382 | 5.19949 | 0.20298 |
|  | 30 | 2000 | 24.0864 | 24.51050 | 15.33264 | 2.91272 |

TABLE 3: The mean fitness values for the Griewank function

| $m$ | Dim. | $G_{max}$ | $SF_0$[5] | $SF_1$ | $DF_2$ | $DF_3$ |
|---|---|---|---|---|---|---|
| 20 | 10 | 1000 | 0.0919 | 0.09609 | 0.08937 | 0.06506 |
|  | 20 | 1500 | 0.0303 | 0.02856 | 0.02863 | 0.02215 |
|  | 30 | 2000 | 0.0182 | 0.01506 | 0.01562 | 0.01793 |
| 40 | 10 | 1000 | 0.0862 | 0.08622 | 0.08170 | 0.05673 |
|  | 20 | 1500 | 0.0286 | 0.02868 | 0.03085 | 0.02150 |
|  | 30 | 2000 | 0.0127 | 0.01348 | 0.01252 | 0.01356 |
| 80 | 10 | 1000 | 0.0760 | 0.07669 | 0.06767 | 0.05266 |
|  | 20 | 1500 | 0.0288 | 0.03109 | 0.02766 | 0.02029 |
|  | 30 | 2000 | 0.0128 | 0.01374 | 0.01345 | 0.01190 |
| 160 | 10 | 1000 | 0.0628 | 0.06373 | 0.06246 | 0.05047 |
|  | 20 | 1500 | 0.0300 | 0.03041 | 0.03145 | 0.01940 |
|  | 30 | 2000 | 0.0127 | 0.01321 | 0.01260 | 0.01029 |

Figure 5 and 6 shows the mean fitness value of the best particle found during 1500 generations with different $w$ and $c_l$ for the Rastrigrin and Griewank function with 20 dimensions, respectively. $c_v$ is fixed as zero. $c_l$ are set as 0 (SPSO) or 0.001 (DPSO). The inertia weights are fixed as 0.4 or linearly decreasing from 0.9 to 0.4, respectively.

For the Rastrigrin function, it can be seen that the performance of DPSO is similar to SPSO during the early stage, however, it will sustainable evolving when the evolution of SPSO is almost stagnated. For the Griewank function, this tendency is weakly but is also exists. For both functions, when $w$=0.4, the performance of the SPSO is the worst; while of the DPSO is the best.

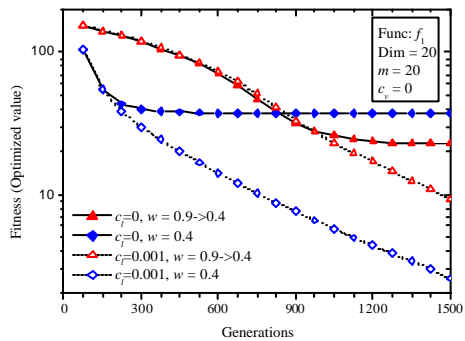

FIG. 5 20-D Rastrigrin function with different $w$ and $c_l$

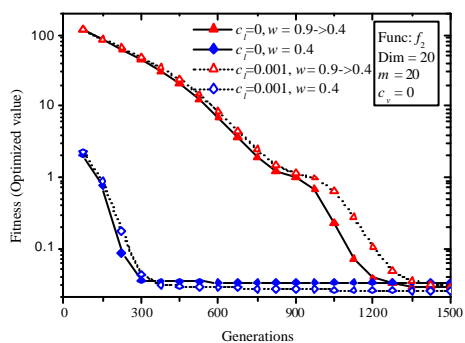

FIG. 6 20-D Griewank function with different $w$ and $c_l$

## 6. Conclusion

Self-organizing has shown extraordinarily flexible and robust in nature systems. By introducing negative entropy into the simple social model of standard PSO through additional chaos, a reasonable open system is constructed, which is in far-from-equilibrium state. With the internal nonlinear interactions among particles, the self-organization of dissipative structure comes into being with the dissipative processes for the introduced negative entropy, which drives the irreversible evolution process toward higher fitness by the selection of keeping best experience. The testing of two multimodal benchmark functions that are commonly used in the evolutionary computation literature indicates the dissipative PSO can improve the performance efficiently.

## References

[1] J. Kennedy, R. Eberhart, " Particle swarm optimization," *Proc. IEEE Int. Conf. on Neural Networks*, pp. 1942-1948, 1995.

[2] R. Eberhart, J. Kennedy, " A new optimizer using particle swarm theory," *Proc. 6th Int. Symposium on Micro Machine and Human Science*, pp. 39-43, 1995.

[3] M. M. Millonas, " Swarms, phase transition, and collective intelligence," In C. G. Langton, (Eds.), *Artificial life III.* Addison-Wesley, MA, 1994.

[4] Y. Shi, R. Eberhart, " Parameter selection in particle swarm optimization," *Proc. 7th Annual Conf. on Evolutionary Programming*, pp. 591-600, 1998.

[5] Y. Shi, R. Eberhart, " Empirical study of particle swarm optimization," *Proc. Congress on Evolutionary Computation*, pp. 1945-1950, 1999.

[6] P. N. Suganthan, " Particle swarm optimiser with neighbourhood operator," *Proc. Congress on Evolutionary Computation*, pp. 1958-1962, 1999.

[7] J. Kennedy, " Stereotyping: improving particle swarm performance with cluster analysis," *Proc. IEEE Int. Conf. on Evolutionary Computation*, pp. 1507-1512, 2000.

[8] R. Eberhart, Y. Shi, " Particle swarm optimization: developments, applications and resources," *Proc. IEEE Int. Conf. on Evolutionary Computation*, pp. 81-86, 2001.

[9] P. J. Angeline, " Evolutionary optimization versus particle swarm optimization: philosophy and performance difference," *Proc. 7th Annual Conf. on Evolutionary Programming*, pp. 601-610, 1998.

[10] I. Prigogine, *Introduction to thermodynamics of irreversible processes.* John Wiley, NY, 1967.

[11] I. Prigogine, " Order through fluctuation: self-organization and social system," In: E. Jantsch (Eds.), *Evolution and consciousness: human systems in transition.* Addison-Wesley, London, 1976.

[12] G. Nicolis, I. Prigogine, *Self-organization in nonequilibrium systems: from dissipative systems to order through fluctuations.* John Wiley, NY, 1977.

[13] H. Haken, *Synergetics.* Springer-Verlag, Berlin, 1983.

[14] E. O. Wilson, *Sociobiology: the new synthesis*, Belknap Press, Cambridge, MA, 1975.

[15] Y. Shi, R. Eberhart, " A modified particle swarm optimizer," *Proc. IEEE Int. Conf. on Evolutionary Computation*, pp. 69-73, 1998.

[16] J. Kennedy, " The particle swarm: social adaptation of knowledge," *Proc. IEEE Int. Conf. on Evolutionary Computation*, pp. 303-308, 1997.

[17] P.M. Allen, *Cities and regions as self-organizing systems: models of complexity.* Gordon and Breach, Amsterdam, 1997.